\pdfoutput=1

\documentclass[11pt]{article}
\usepackage{placeins}

\usepackage[final]{acl}
\usepackage{tcolorbox}
\usepackage{times}
\usepackage{latexsym}
\usepackage{booktabs} 
\usepackage[T1]{fontenc}

\usepackage{listings}
\usepackage[utf8]{inputenc}
\usepackage{multirow}

\usepackage{inconsolata}

\usepackage{graphicx}

\usepackage{multicol}

\usepackage{kotex}

\title{Secure Multifaceted-RAG for Enterprise: Hybrid Knowledge Retrieval with Security Filtering}

\author{
\textbf{Grace Byun}\textsuperscript{1},
  \textbf{Shinsun Lee}\thanks{Work done while at Emory University as a visiting scholar.}\textbf{\textsuperscript{1,2}}, 
  \textbf{Nayoung Choi}\textsuperscript{1},   
\textbf{Jinho D. Choi\textsuperscript{1}} \\
  \textsuperscript{1}Emory University, \textsuperscript{2}Hyundai Motor Company \\
  \small{\texttt{\{grace.byun, shinsun.lee, 
 nayoung.choi,  jinho.choi\}@emory.edu}}
}

\begin{document}
\maketitle
\begin{abstract}
Existing Retrieval-Augmented Generation (RAG) systems face challenges in enterprise settings due to limited retrieval scope and data security risks. When relevant internal documents are unavailable, the system struggles to generate accurate and complete responses. Additionally, using closed-source Large Language Models (LLMs) raises concerns about exposing proprietary information. To address these issues, we propose the \textbf{Secure Multifaceted-RAG (SecMulti-RAG)} framework, which retrieves not only from internal documents but also from two supplementary sources: pre-generated expert knowledge for anticipated queries and on-demand external LLM-generated knowledge. To mitigate security risks, we adopt a local open-source generator and selectively utilize external LLMs only when prompts are deemed safe by a filtering mechanism. This approach enhances completeness, prevents data leakage, and reduces costs. In our evaluation on a report generation task in automotive industry, SecMulti-RAG significantly outperforms traditional RAG—achieving 79.3–91.9\% win rates across correctness, richness, and helpfulness in LLM-based evaluation, and 56.3–70.4\% in human evaluation. This highlights SecMulti-RAG as a practical and secure solution for enterprise RAG.

\end{abstract}

\section{Introduction}
Retrieval-Augmented Generation (RAG) \citep{rag} has become a powerful tool for AI-driven content generation. However, existing RAG frameworks face significant limitations in enterprise applications. Traditional RAG systems rely heavily on internal document retrieval, which can lead to incomplete or inaccurate responses when relevant information is missing. Moreover, leveraging external Large Language Models (LLMs) like GPT \citep{openai2024gpt4technicalreport}, Claude \citep{TheC3}, or DeepSeek \citep{deepseekai2025deepseekr1incentivizingreasoningcapability} introduces security risks and high operational costs, making them less viable for enterprise deployment.

To address these challenges, we introduce SecMulti-RAG framework that optimizes information retrieval, security, and cost efficiency. Our approach integrates three distinct sources: (1) dynamically updated enterprise knowledge base, (2) pre-written expert knowledge for anticipated queries, and (3) on-demand external knowledge, selectively retrieved when user prompt is safe. For sercurity, we introduce a filtering mechanism that ensures proprietary corporate data is not sent to external models. Furthermore, instead of relying on powerful closed-source LLMs, we use a local open-source model as the primary generator, selectively invoking external models only when user prompts are non-sensitive.

In this paper, we apply SecMulti-RAG to the Korean automotive industry. Our fine-tuned filter, retriever, and generation models show strong performance, ensuring the reliability of our approach. On the report generation task, our method outperforms the traditional RAG approach in correctness, richness, and helpfulness, as evaluated by both humans and LLMs. We also present adaptable strategies to meet specific environmental needs, emphasizing its flexibility. Our key contributions are:

\begin{itemize}
    \item Multi-source RAG framework combining internal knowledge, pre-written expert knowledge, and external LLMs to enhance response completeness.
    \item Confidentiality-preserving filter to prevent exposure of sensitive corporate data to external LLMs.
    \item Cost-efficient approach that leverages high-quality retrieval to compensate for smaller local generation models.
\end{itemize}

\section{Related Work}
\paragraph{Enhancing Retrieval-Augmented Generation}
Many efforts have been made to enhance RAG systems \citep{zhou2024assistragboostingpotentiallarge, neuro}.
\citet{jeong2024adaptiveraglearningadaptretrievalaugmented} classify user prompts based on complexity to determine the optimal retrieval strategy, making their approach relevant to our filtering mechanism for selecting retrieval sources.
Meanwhile, \citet{yu2023generateretrievelargelanguage} and \citet{wu2024multisourceretrievalquestionanswering} replace traditional document retrieval with generative models. In particular, \citet{wu2024multisourceretrievalquestionanswering} propose a multi-source RAG (MSRAG) framework that integrates GPT-3.5 with web-based search, making it the most relevant to our work. In contrast, our approach retains internal document retrieval while integrating pre-generated expert knowledge and external LLMs. 

\paragraph{Security Risks in LLM}

As generative models are widely used, concerns about security and privacy risks continue to grow. Many studies have explored methods for detecting and mitigating the leakage of sensitive information \citep{zhang2024ghostpastidentifyingresolving, kim2023propileprobingprivacyleakage,Hayes2017LOGANEP,lukas2023analyzingleakagepersonallyidentifiable}. For example, \citet{chong2024casperpromptsanitizationprotecting} present a prompt sanitization technique that enhances user privacy by identifying and removing sensitive information from user inputs before they are processed by LLM services. Our study incorporates a user prompt filtering mechanism, ensuring a more secure retrieval process.

\section{Method}

\begin{figure}[ht]
  \centering
  \includegraphics[width=0.85\columnwidth]{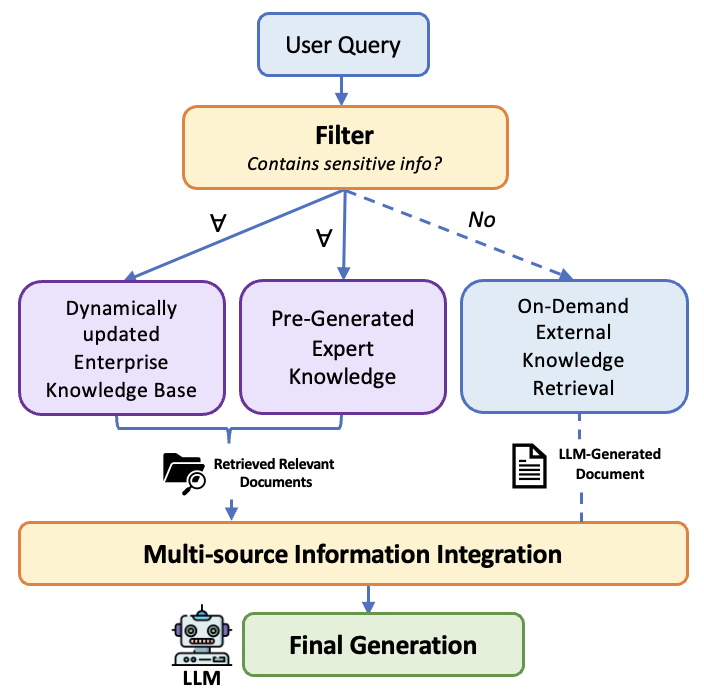}
  \caption{SecMulti-RAG framework}
  \label{fig:rag_framework}
\end{figure}

\noindent As shown in Figure \ref{fig:rag_framework}, our RAG framework consists of three core components: multi-source retrieval, a confidentiality-preserving filtering mechanism, and local model adaptation. (Appendix \ref{sec:appendix_framework})

\paragraph{1) Multi-Source Retrieval}  
Unlike conventional RAG frameworks that rely solely on internal structured document chunks, our system retrieves information from three distinct sources: (1) internal corporate documents, (2) pre-generated high-quality answers to anticipated queries, and (3) real-time external knowledge generated by closed-source LLMs. This multi-source retrieval strategy improves response completeness and accuracy, especially when internal documents are insufficient.

\paragraph{2) Confidentiality-Preserving Filtering Mechanism}  
To mitigate the risk of unintended data leakage when interacting with external closed-source LLMs, we introduce a query filtering mechanism that detects security-sensitive content. If a user query is classified as containing confidential information, external retrieval is skipped, and the system generates responses solely based on internal documents and pre-curated expert knowledge. This mechanism ensures data confidentiality while maintaining retrieval quality (Section \ref{sec:filter}).

\paragraph{3) Local Model Adaptation}  


Since enterprises often encounter security and cost limitations when using powerful closed-source LLMs, we use open-source Qwen-2.5-14B-Instruct \cite{qwen} as our primary generation model. We fine-tune this model using domain-specific data to better reflect the language and knowledge of the Korean automotive domain. (Section \ref{sec:model})

\paragraph{4) End-to-End RAG Pipeline}  
The final system consists of a multi-stage pipeline where user queries are processed through filtering, retrieval, and generation stages. By integrating high-quality retrieval with local model adaptation, our framework demonstrates that a well-optimized retrieval system can compensate for the limitations of smaller, locally deployed LLMs, making enterprise RAG both scalable and secure.

\begin{table*}[ht]
    \centering\small
    \begin{tabular}{p{7cm}|c|p{3.4cm}|p{3.2cm}}
        \toprule
        \textbf{Query} & \textbf{Label} & \textbf{Reason} & \textbf{Type}\\
        \hline
        {What is the reason behind the introduction of the IIHS small overlap crash test, and how has it influenced vehicle design?} & \textcolor{blue}{1} & A general question about publicly available test standards and their impact. & General query \\
        \hline
        {What is the deformation value in the second-row passenger compartment due to the upper bending near the chassis frame fuel tank MTG during HD3 52kph rear evaluation?} & \textcolor{red}{0} & Contains detailed information about structural vulnerabilities. & Security-sensitive query with project names \\
        \hline
        {Why is it necessary to change the location of the door pushing bracket, and what is the problem with the current location?} &\textcolor{red}{0}  & Reveals structural design weaknesses. &Security-sensitive query without project names \\
        \bottomrule
    \end{tabular}
    \vspace{-0.5em}
    \caption{Examples of safe and unsafe queries and their corresponding reasons, translated into English. Label 0 refers to security-sensitive queries, while 1 refers to non-sensitive ones. (Specific vehicle type is anonymized.)}
    \label{tab:filter_query_exmple}
\end{table*}

\section{Confidentiality Filter} \label{sec:filter}

\subsection{Dataset}

\noindent Security-sensitive and general (non-sensitive) queries were created by Korean automotive engineers with the assistance of the Claude 3.7 Sonnet \citep{TheC3}, accessed through a university-internal service built on AWS Bedrock\footnote{\url{https://aws.amazon.com/ko/bedrock/}}. This setup provides secure access to Claude without exposing data to external LLM providers\footnote{All Claude 3.7 Sonnet model used in this study—for data generation and LLM-based evaluation—were accessed exclusively through this secure university-internal service.}.  Figure~\ref{fig:query_prompt} in Appendix~\ref{sec:appendix_query} illustrates the prompt template used to construct the query dataset. The prompts are designed to elicit three types of queries: (1) \textbf{general queries} that do not pose confidentiality risks, (2) \textbf{security-sensitive queries (easy)} containing explicit project names, and (3) \textbf{security-sensitive queries (hard)} that omit project names. Type (3) queries are particularly challenging to classify, even for expert engineers, due to the absence of clear identifiers. In addition to the queries and the binary labels (sensitive or non-sensitive), brief rationales are generated to explain the reasoning behind each label. Table~\ref{tab:filter_query_exmple} presents example queries, and Table~\ref{tab:split_size_version1} summarizes data statistics.

\begin{table}[htbp]
\centering
\begin{tabular}{lcccc}
\toprule
\textbf{Set}  & \textbf{Safe} & \multicolumn{2}{c}{\textbf{Unsafe}} & \textbf{Total}\\
\cmidrule(lr){3-4}
& &  \textbf{Easy} & \textbf{Hard}& \\
\midrule
Train       & 820 & 800 & 720 & 2,340\\
Validation   & 102 & 100 & 90 & 292 \\
Test        & 103 & 101 & 90 & 294  \\
\bottomrule
\end{tabular}
\vspace{-0.5em}
\caption{Data split for training and evaluation of the filter model.}
\label{tab:split_size_version1}
\end{table}

\subsection{Model}
For the filter, we fine-tune a lightweight model, Qwen2.5-3B-Instruct \cite{qwen}, to classify safe and unsafe queries. The training parameters and the prompt used for the filtering process are detailed in Appendix \ref{sec:appendix_filter}.

\subsection{Evaluation}

\noindent As shown in Table~\ref{tab:filter-eval_version1}, we evaluate the filter model on two subsets: \textbf{Easy-only} (queries with explicit project names) and \textbf{Easy\&Hard} (both with and without project names). Security-sensitive queries (class 0) are treated as the positive class, making recall critical in preventing information leakage to external LLMs. For easy test cases, the filter achieves 99.01\% recall, indicating strong performance with minimal false negatives. When ambiguous queries is included, accuracy drops to 82.31\% and recall to 74.35\%, while precision remains high (97.93\%), meaning that when it does flag a query as sensitive, it is highly likely to be correct. To estimate the upper bound, we conduct human evaluation on the \textbf{Easy\&Hard} set. Expert annotator achieves 80.95\% accuracy, 92.99\% precision, and 76.44\% recall, highlighting the intrinsic ambiguity and difficulty of the task.

\begin{table}[htbp]
\centering
\begin{tabular}{llccc}
\toprule
\textbf{Method} & \textbf{Test Data} & \textbf{Acc \%} & \textbf{Prec} & \textbf{Rec} \\
\midrule
\multirow{2}{*}{\textbf{Filter}} 
    & \textbf{Easy-only} & 98.04 & 97.09 & 99.01 \\
    & \textbf{Easy\&Hard} & 82.31 & 97.93 & 74.35 \\
\midrule
\textbf{Human} & \textbf{Easy\&Hard} & 80.95 & 92.99 & 76.44 \\
\bottomrule
\end{tabular}
\vspace{-0.5em}
\caption{Evaluation results of the filter model and human annotators on the test set. \textbf{Easy-only} includes security-sensitive queries with explicit project names. \textbf{Easy\&Hard} includes both easy (with project names) and hard (without project names) queries. \textbf{Human} shows expert-labeled upper bound performance.}
\label{tab:filter-eval_version1}
\end{table}

\subsection{Application}
In practical deployment, constructing a labeled dataset of safe and unsafe queries for filter training can be labor-intensive. To address this, we propose a progressive deployment strategy for the confidentiality filter. Initially, the system operates without filtering or external retrieval, relying only on internal documents and pre-generated expert knowledge. During this phase, real user queries are collected and later labeled to train a filter model, enabling cost-effective integration over time. Alternatively, the filter can perform query rewriting—flagged queries are transformed into safer versions, allowing secure forwarding to external LLMs. This flexible design supports scalable adaptation to organizational privacy and deployment needs.

\section{Retrieval}

\noindent This paper focuses on generating reports for enterprise-level engineering problems. Our retrieval system is built upon 6,165 chunked documents, consisting of 5,625 chunks from the Enterprise Knowledge Base and 540 from Pre-written Expert Knowledge. To prevent data leakage, only the training subset of the 675 Pre-written Expert Knowledge documents is included in the retrieval pool; the remaining 135 keyword–report pairs are reserved for the final evaluation of \textbf{SecMulti-RAG} (Section~\ref{sec:final_evaluation}). All documents listed in Table~\ref{tab:dataset_document} are indexed using FAISS \citep{douze2025faisslibrary}, and our trained retriever retrieves the top five most relevant documents for each query.

\begin{table}[htbp]
\centering
\small
\resizebox{\columnwidth}{!}{%
\begin{tabular}{p{3.95cm} p{2cm} c c}
\toprule
\textbf{Type} & \textbf{Source} & \textbf{File/Page} & \textbf{Chunks} \\
\midrule
\multirow{3}{*}{\textbf{{Enterprise Knowledge Base}}}
 & Test Report    & 1{,}463 & 4{,}662 \\
 & Meeting Report & 249     & 882    \\
 & Textbook       & 404     & 81     \\
\midrule
\multirow{1}{*}{\textbf{{Pre-written Expert Knowledge}}}
 & \parbox{4.5cm}{Gold Report} & {-} & 540 \\
\midrule
\multicolumn{2}{l}{\textbf{Total}} & {} & \textbf{6{,}165} \\
\bottomrule
\end{tabular}%
}
\vspace{-0.5em}
\caption{Overview of chunked documents used in SecMulti-RAG retrieval. Traditional RAG retrieves only from the Enterprise Knowledge Base.}
\label{tab:dataset_document}
\end{table}

\subsection{Dataset}
\subsubsection{Enterprise Knowledge Base}\label{sec:qa_data}
For the Enterprise Knowledge Base dataset, we use the dataset introduced by \citet{choi2025trustworthyanswersmessierdata}. It consists of test reports, meeting reports, and textbooks, with each document segmented into meaningful units such as slides, chapters, or other relevant sections. QA pairs from the reports and textbook are used to train both retriever and generation model. See Table \ref{tab:dataset_split} in Appendix \ref{sec:appendix_ekb_stat} for details.

\subsubsection{Pre-written Expert Knowledge}\label{sec:prewritten_expert_data}
To construct a high-quality knowledge source, a domain expert in automotive engineering first curates a list of domain-specific keywords, representing expertise-level problems. Using these keywords, the expert then generates pre-written expert knowledge using Claude (Appendix~\ref{sec:appendix_keyword}, \ref{sec:appendix_prewritten_report_generation}). A total of 675 keyword–report pairs are partitioned into training, validation, and test splits in an 8:1:1 ratio.


\subsubsection{External Knowledge from LLM}

After the user query passes the safety filter and is deemed safe, we use GPT-4o\footnote{\url{https://platform.openai.com/docs/models/gpt-4o}} to provide on-demand external knowledge. Specifically, in this paper, it generates a general-purpose technical background document to assist engineers in drafting formal safety reports. The prompt is illustrated in Appendix \ref{sec:appendix_external}. The generated document is then indexed into the document pool for future retrieval.

\subsection{Retriever} \label{sec:retriever}
We fine-tune BGE-M3 \cite{chen-etal-2024-m3}, a multilingual encoder supporting Korean, using QA and keyword–report pairs (Sections~\ref{sec:qa_data}, \ref{sec:prewritten_expert_data}). Training is done for 10 epochs on 4 × 48GB RTX A6000 GPUs with publicily available code\footnote{\url{https://github.com/FlagOpen/FlagEmbedding}}. For evaluation, we use all splits (training, validation, and test) as the chunk pool to ensure sufficient data coverage and mitigate potential biases due to the small size of the test set.

\subsection{Retriever Evaluation}

\noindent The performance of the retriever is evaluated using Mean Average Precision (MAP@k), which calculates the average precision of relevant results up to rank $k$. As shown in Table \ref{tab:retriever_performance}, fine-tuning BGE-M3 leads to significant improvements, underscoring the importance of task-specific adaptation.

\begin{table}[htbp]
\centering
\small
\begin{tabular}{c|rrr}
\toprule
\textbf{Model} & \multicolumn{1}{c}{\textbf{MAP@1}} & \multicolumn{1}{c}{\textbf{MAP@5}} & \multicolumn{1}{c}{\textbf{MAP@10}} \\ \midrule
BGE (Vanilla)    & 0.2855 & 0.3793 & 0.3925 \\
BGE (Fine-tuned)  & \textbf{0.5965} & \textbf{0.7027} & \textbf{0.7099} \\ \bottomrule
\end{tabular}
\vspace{-0.4em}
\caption{Comparison of retrieval performance between the vanilla and fine-tuned models on our test dataset.}
\label{tab:retriever_performance}
\vspace{-1em}
\end{table}

\subsection{Document Selection Strategy} \label{sec:selection}

In this study, we rank candidate documents by semantic similarity and apply a selection constraint: at most one external knowledge is included per query. GPT-generated documents are limited to one per query as they provide only general technical background, and excessive reliance on such external content may reduce the factual grounding of responses. Although this constraint is currently implemented via heuristic rules, we aim to develop a learning-based document selection strategy that jointly considers query characteristics and document provenance.


\section{Generation}  \label{sec:model}

\subsection{Generator}

We use Qwen-2.5-14B-Instruct \cite{qwen} as our base language model, as it is one of the few multilingual models that officially support Korean while offering a sufficient context length. We fine-tune the model using QA pairs introduced in Section \ref{sec:qa_data}. Hyperparameters, GPU configurations, and the generation prompt are in Appendix~\ref{sec:appendix_qwen_train}.

\subsection{Result}

Table~\ref{tab:generation_summary} summarizes the retrieved document sources and filtering results for the 135 test queries. All queries retrieved pre-generated expert knowledge, while 28.1\% and 18.5\% also retrieved GPT-generated and internal documents, respectively. Among the test queries, 34 are classified as safe, enabling on-demand GPT generation. Interestingly, the number of queries retrieving GPT-generated documents slightly exceeds the number of safe queries. This is because previously generated external documents remain in the retrieval pool and can still be retrieved for relevant future queries, even if those queries are classified as sensitive. This illustrates a key benefit of our system: as more external knowledge is accumulated over time, the retrieval pool becomes richer, allowing even sensitive queries to benefit from external knowledge without compromising security.

\begin{table}[ht]
\centering
\begin{tabular}{l r}
\toprule
\textbf{Category} & \textbf{Count (\%)} \\
\midrule
\textbf{Pre-written knowledge} retrieved & 135 (100\%)\\
\textbf{External knowledge} retrieved & 38 (28.1\%)\\
\textbf{Internal document} retrieved & 25 (18.5\%)\\
\midrule
\textbf{filter=1} (Safe) & 34 (25.2\%)\\ 
\textbf{filter=0} (Security-sensitive)  & 101 (74.8\%)\\
\bottomrule
\end{tabular}
\vspace{-0.4em}
\caption{Distribution of retrieved document types (appearing at least once among the Top-5 documents) and filtering outcomes in the SecMulti-RAG}
\label{tab:generation_summary}
\end{table}

\section{Evaluation}\label{sec:final_evaluation}

\subsection{Method}
To evaluate the effectiveness of our approach, we conduct a qualitative assessment based on three metrics: Correctness, Richness, and Helpfulness. We perform pairwise comparisons using both LLM-as-a-judge \citep{zheng2023judgingllmasajudgemtbenchchatbot} and human evaluation, comparing responses generated by \textbf{Traditional RAG}—which retrieves only from the internal knowledge base—with our \textbf{SecMulti-RAG}, which retrieves from the internal knowledge base, pre-generated expert knowledge, and on-demand external knowledge. For each metric, Claude and a human annotator assess which response (A or B) is better and record the outcome as a win, loss, or tie. To mitigate position bias from the judge LLM, we anonymize the response order by randomly assigning either SecMulti-RAG or Traditional RAG as response A or B in half of the cases. The prompt that we use for LLM-based evaluation is in Appendix \ref{sec:appendix_llm_judge_prompt}.

\vspace{-0.5em}
\begin{itemize}
\setlength\itemsep{-0.05em}
    \item \textbf{Correctness} assesses the factual consistency with the given gold answer. The pre-written reports are provided as gold answers.
    \item \textbf{Richness} evaluates the level of detail and completeness in the response.
    \item \textbf{Helpfulness} measures how clear, informative, and useful the response is.
\end{itemize}


\subsection{Result}

\begin{figure}[htbp]
\centering
\includegraphics[width=\columnwidth]{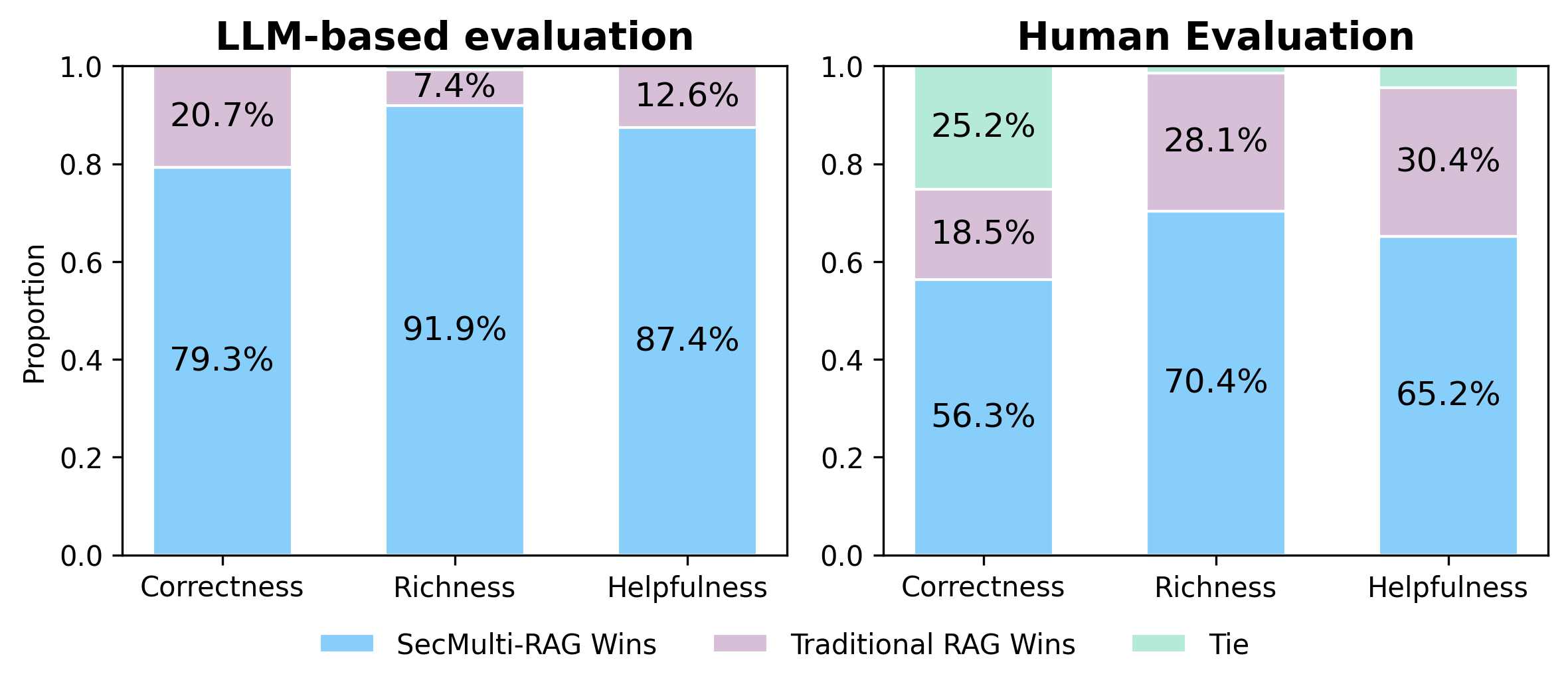}
\vspace{-1.2em}
\caption{Win rate comparison between SecMulti-RAG and Traditional RAG across evaluation metrics}
\label{fig:pie_multisource_vs_traditional}
\vspace{-0.5em}
\end{figure}

\noindent Figure \ref{fig:pie_multisource_vs_traditional} presents a comparison of win rates between the two systems, evaluated by human annotators and an LLM-based evaluation. Both evaluation sources consistently prefer the outputs of SecMulti-RAG, particularly in the richness metric. There is a substantial agreement between human and LLM evaluations as in Appendix \ref{sec:appendix_agreement}.


\begin{figure*}[ht]
  \centering
\includegraphics[width=2.05\columnwidth]{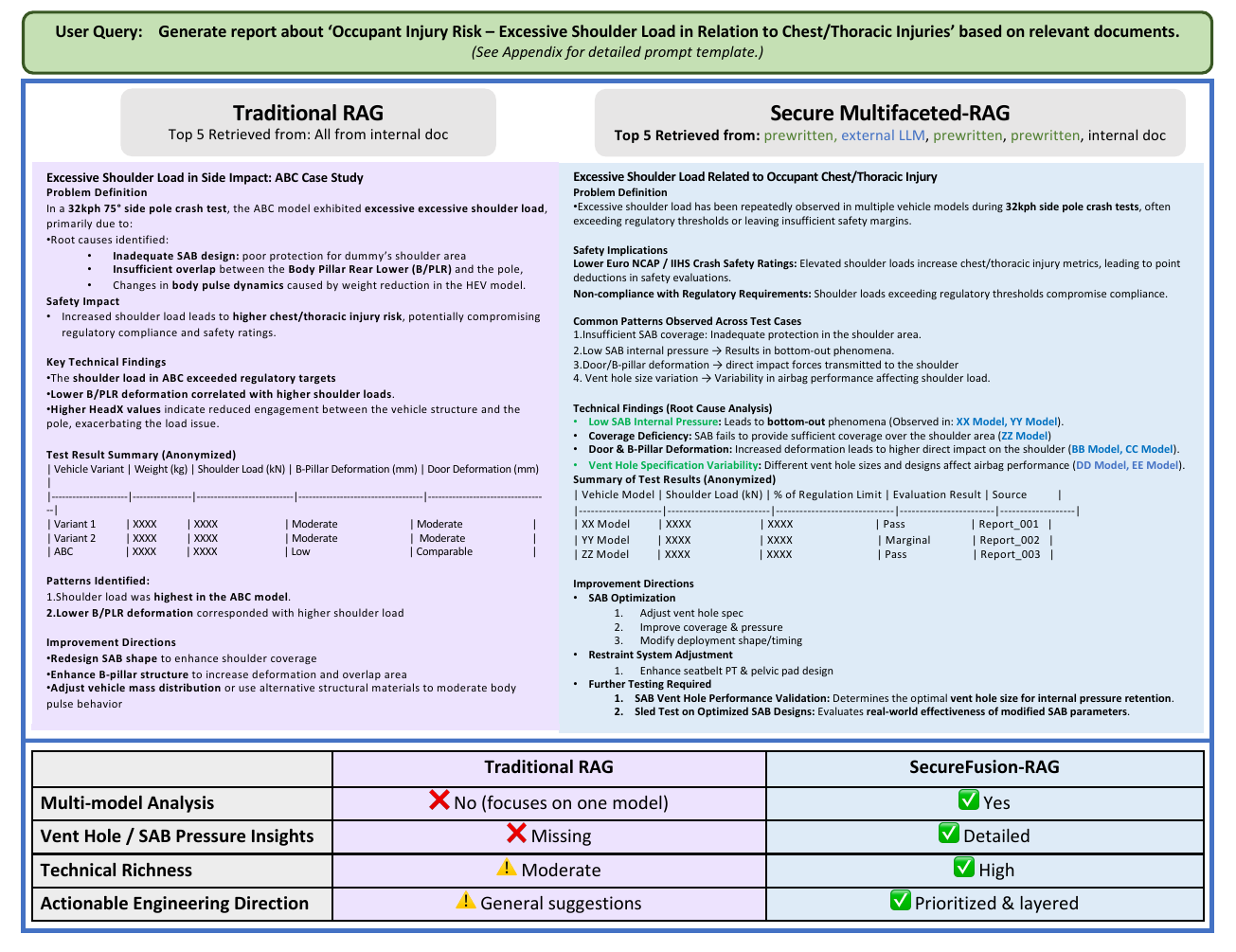}
\vspace{-0.8em}
  \caption{Comparison Between Traditional RAG and SecMulti-RAG (Translated from Korean; Sensitive Information Anonymized)}
  \label{fig:example_report_rag}
\end{figure*}

\subsection{Analysis}
While human evaluators tend to assign `tie' labels more frequently than the judge LLM, both clearly favor SecMulti-RAG across all metrics. Figure \ref{fig:example_report_rag} illustrates an example of reports generated by SecMulti-RAG and Traditional RAG. 

\noindent As expected, richness is the most notably improved aspect of our framework, reflecting the benefit of incorporating diverse documents from multiple sources. SecMulti-RAG outputs contain significantly more detailed information, such as more diverse test cases and technical findings. In fact, the average length of reports generated by SecMulti-RAG is 2,660.21 tokens, compared to 1,631.84 tokens from Traditional RAG. In terms of correctness, both systems generally produce factually accurate content but occasionally fail to cite the correct source document. For helpfulness, while our approach delivers more comprehensive reports, Traditional RAG may be more favorable when the engineer’s intent is to focus on a specific test case, as its responses tend to be more narrowly scoped. This could potentially be mitigated through prompt tuning. One notable issue is the occasional generation of Chinese characters, due to the Qwen model's Chinese-centric pretraining. This problem is likely caused by the longer documents retrieved by SecMulti-RAG, which increase the context length and make the model more vulnerable to generating such errors.

\section{Conclusion}

In this paper, we present SecMulti-RAG for enterprise that integrates internal knowledge bases, pre-generated expert knowledge, and on-demand external knowledge. Our framework introduces a confidentiality-aware filtering mechanism that protects security-sensitive user prompts by bypassing external augmentation when necessary, mitigating the risk of information leakage to closed-source LLMs. In our experiments on automotive engineering report generation, SecMulti-RAG showed clear improvements over Traditional RAG in terms of correctness, richness, and helpfulness.  It achieved win rates ranging from 56.3\% to 70.4\%, as evaluated by human evaluators, outperforming traditional RAG across all metrics. Beyond performance gains, our approch is a cost-efficient, privacy-preserving, and scalable solution, leveraging high-quality retrieval with locally hosted LLMs.

\section*{Limitations}


Due to the lack of publicly available Korean-language datasets in the automotive domain, our evaluation is limited to the report generation task based on a relatively small amount of data that we have constructed ourselves. While this work is intended for an industry track and demonstrates practical significance, future research could enhance the academic impact by showing the scalability of the SecMulti-RAG framework across a broader range of tasks and domains. In fact, we have conducted preliminary experiments on engineering question answering tasks beyond report generation, and observed that SecMulti-RAG also performs well in those scenarios, indicating its potential. Example responses are provided in Appendix \ref{sec:appendix_qa}.

Currently, we apply a heuristic selection constraint: at most one external knowledge document is included per query (Section \ref{sec:selection}). This constraint is motivated by our observation that GPT-generated documents, while useful for providing general technical background or engineering context, may dilute the factual grounding of system responses if overused. However, this rule-based constraint lacks adaptability and may not always yield optimal document combinations tailored to diverse query intents. In future work, we aim to develop a learning-based document selection strategy that jointly considers query characteristics and document sources, allowing the system to automatically re-rank and select optimal document combinations to better match the specific needs of each query and deployment context.

\section*{Acknowledgement}
We gratefully acknowledge the support of the Hyundai Motor. Any opinions, findings, and conclusions or recommendations expressed in this material are those of the authors and do not necessarily reflect the views of Hyundai Motor.

\bibliography{custom}

\newpage

\appendix

\section{RAG Framework} \label{sec:appendix_framework}

\begin{figure*}[htbp]
  \centering
  \includegraphics[width=1.3\columnwidth]{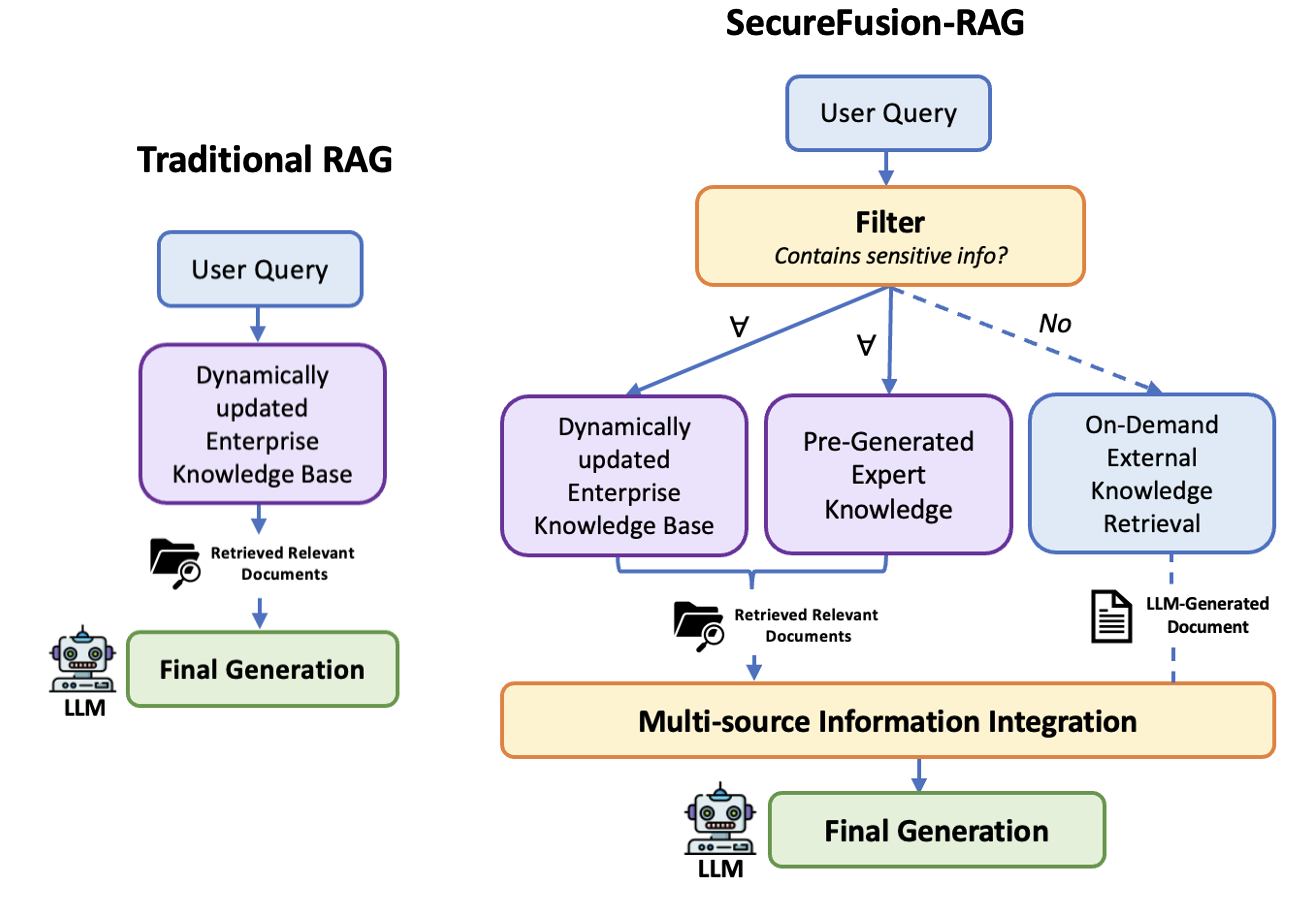}
  \caption{Traditional RAG vs SecMulti-RAG}
  \label{fig:framework_compare}
\end{figure*}

Figure \ref{fig:framework_compare} demonstrates both Traditional RAG and SecMulti-RAG framework.

\section{Filter}
\label{sec:appendix_filter}

\subsection{Query Dataset} \label{sec:appendix_query}

Security-sensitive and general (non-sensitive) queries are carefully constructed by Korean automotive engineers with the assistance of Claude. Figure \ref{fig:query_prompt} shows the prompt template that the engineers used during data construction, reflecting our efforts to optimize prompt design for query generation.

\begin{figure*}[!htbp]
    \tiny
    \begin{tcolorbox}[width=\textwidth, colback=white, colframe=black, title=Prompt, sharp corners]
\begin{lstlisting}[
            language=,
            basicstyle=\ttfamily\tiny,
            breaklines=true,
            frame=none
        ]
[Security-sensitive queries without project name]

You are an engineer in charge of vehicle crash safety testing. Based on internal reports, you are tasked with generating sensitive security-related queries that should not be exposed to the public. The queries must meet the following requirements:

Query Generation Guidelines:
- Do NOT include any specific project names or test IDs.
- Do NOT ask about exact internal figures or detailed designs directly. Instead, formulate questions that indirectly imply such information through the context.
- Each question should be capable of revealing internal issues or technical weaknesses if exposed externally.
- Vary the length of the queries (mostly short questions, but some longer ones included).
- Generate 50 questions in total.
- Provide a one-line explanation for why each question is considered security-sensitive.

Example Questions:
- "What is the root cause of the excessive B-pillar deformation observed in recent tests?" (Indirectly reveals an internal issue)
- "What causes the delayed airbag deployment timing?" (Hints at internal system problems)
- "Why is reinforcement needed in the door hinge area?" (Implies structural vulnerability)
- "Why does a specific area repeatedly exhibit excessive deformation in crash tests?" (Reveals unresolved structural weaknesses)

Please refer to the examples above and generate 50 diverse security-sensitive questions that indirectly reveal sensitive internal information.

Here are the internal reports:
<Reports>
{reports}
</Reports>


=======================================================================================================================
[Security-sensitive queries with project name]

You are an engineer in charge of vehicle crash safety testing. Based on internal reports, you are tasked with generating security-related queries that include a specific project name but exclude test IDs. The queries must meet the following requirements:

Query Generation Guidelines:
1. MUST include a specific project name (but do NOT include test IDs).
2. In most queries, place the project name at the beginning of the sentence, but vary the position in some queries (middle or end).
3. It is acceptable to include sensitive internal figures or specific design information.
4. The queries must be security-sensitive and unsuitable for public exposure.
5. Vary the length of the queries (mostly short, but some longer ones included).
6. Provide a one-line explanation for why each question is considered security-sensitive.
7. Maintain the following consistent output format:

- "Query sentence"  
(Explanation)

Here are the internal reports:
<Reports>
{reports}
</Reports>


=======================================================================================================================
[General (non-sensitive) queries]

You are an engineer in charge of vehicle crash safety testing. Based on internal reports, generate general queries that do not pose any security concerns and can be disclosed publicly. These queries should reflect realistic engineering questions that professionals may encounter in practice. The queries must meet the following requirements:

Query Generation Guidelines:
1. Do NOT include any specific project names or test IDs.
2. Avoid questions about sensitive internal figures or detailed designs. Instead, focus on general technical concepts, practical engineering concerns, and issues that engineers would typically discuss.
3. Questions should be centered on issues arising in crash safety testing environments, engineering decision-making, and practical field concerns.
4. Vary the length of the queries (mostly short, but some longer ones included).
5. Provide a one-line explanation for why each question is NOT considered security-sensitive.
6. Maintain the following consistent output format:

- "Query sentence"  
(Explanation)

Here are the internal reports:
<Reports>
{reports}
</Reports>
\end{lstlisting}
    \end{tcolorbox}
    \caption{Prompt used for query data generation (Translated to English)}
    \label{fig:query_prompt}
\end{figure*}

\subsection{Training}
Full fine-tuning is performed on Qwen2.5-3B-Instruct \citep{qwen} using a per-device batch size of 2 and a gradient accumulation step of 256, effectively simulating a large batch size. The model is trained for 3 epochs with a learning rate of 8e-6. Training is conducted on 3 $\times$ 48GB RTX A6000 GPUs.

\subsection{Prompt}
Figure \ref{fig:filter_prompt} provides the prompt that we use to filter out user queries that should not be exposed to closed-source LLM.

\begin{figure*}[ht!]
    \tiny
    \begin{tcolorbox}[width=\textwidth, colback=white, colframe=black, title=Prompt, sharp corners]
\begin{verbatim}
You are a classification model for user queries.
There are **TWO CLASSES ONLY**:
0 => The query may contain internal or sensitive information and is NOT safe.
1 => The query is general and can be safely processed.

**IMPORTANT INSTRUCTION**:
- If the query contains any internal project names, testing scores, engineering design details, 
weakness (defects), and internal solution, you MUST classify as 0.
- If uncertain, choose 0.

You must output ONLY one number: either 0 or 1.

Examples:

    [Example 1] (GPT Not Allowed - Corporate Internal Information)
    User Question: "Tell me the test results of LX2 P2."
    Judgment: 0
    
    [Example 2] (GPT Not Allowed - Requesting Internal Evaluation Report for a Specific Vehicle)
    User Question: "Share the internal crash test report of the G70 frontal crash test."
    Judgment: 0
    
    [Example 3] (GPT Allowed - General Automotive Safety Regulations)
    User Question: "Explain the U.S. Federal Motor Vehicle Safety Standard."
    Judgment: 1
    
    [Example 4] (GPT Allowed - Structural Improvement Cases for Automotive Components)
    User Question: "Tell me about cases of subframe failure and possible improvements."
    Judgment: 1
    
    [Example 5] (GPT Not Allowed - Confidential Design Information of a Specific Company)
    User Question: "Tell me the key design changes in Hyundai's new engine blueprint."
    Judgment: 0
    
    [Example 6] (GPT Allowed - General Mechanical Engineering Theory)
    User Question: "What are some methods to increase the rigidity of a car chassis?"
    Judgment: 1
    
    [Example 7] (GPT Not Allowed - Requesting Internal Test Data)
    User Question: "Tell me the results of the in-house crash tests conducted by HMG."
    Judgment: 0
    
    [Example 8] (GPT Allowed - Public Data-Based Information)
    User Question: "Explain the European NCAP crash test standards and evaluation criteria."
    Judgment: 1



\end{verbatim}
    \end{tcolorbox}
\caption{Prompt used for Filtering Progress (Translated to English)}
\label{fig:filter_prompt}
\end{figure*}

\section{Enterprise Knowledge Base} \label{sec:appendix_ekb_stat}

Table \ref{tab:dataset_split} shows the statistics of the Enterprise Knowledge Base dataset. Document chunks and QA pairs are constructed from test reports, meeting reports, and textbook. Refer to \citet{choi2025trustworthyanswersmessierdata} for more details.

\begin{table}[htbp]
\resizebox{\columnwidth}{!}{%
\begin{tabular}{l|l|rrr|r}
\toprule
\multicolumn{1}{c|}{\textbf{Source}} & \multicolumn{1}{c|}{\textbf{Data}} & \multicolumn{1}{c}{\textbf{Train}} & \multicolumn{1}{c}{\textbf{Val}} & \multicolumn{1}{c|}{\textbf{Test}} & \multicolumn{1}{l}{\textbf{Total}} \\ \midrule
\multirow{2}{*}{Test Report}         & Chunk                              & 3,729                              & 466                              & 467                                & 4,662                              \\
                                     & QA Pair                          & 47,660                             & 5,823                            & 5,919                              & 59,402                             \\ \midrule
\multirow{2}{*}{Meeting Report}      & Chunk                              & 705                                & 88                               & 89                                 & 882                                \\
                                     & QA Pair                          & 6,144                              & 752                              & 800                                & 7,696                              \\ \midrule
\multirow{2}{*}{Textbook}            & Chunk                              & 64                                 & 8                                & 9                                  & 81                                 \\
                                     & QA Pair                          & 1,182                              & 162                              & 161                                & 1,505                              \\ \bottomrule
\end{tabular}%
}
\caption{Enterprise Knowledge Base Dataset: Statistics by source, detailing the distribution of chunks and QA pairs.}
\label{tab:dataset_split}
\end{table}

\section{Pre-written Expert Knowledge}
\subsection{Keywords} \label{sec:appendix_keyword}

Below are some of the keywords we use to generate pre-written expert knowledge. Reports related to these keywords (main problems) are pre-generated by automotive engineers.

\textbf{1. 차체 구조 및 안전성 관련 이슈 (Vehicle Structural Integrity and Safety Issues)}

\noindent
\textbf{1. 차체 구조 및 구조적 완전성 (Body Structure and Structural Integrity)}

\noindent
\textbf{1.1. 필러(Pillar) 관련 문제 (Pillar-Related Issues)}

\noindent
\textbf{1.1.1. A필러 문제 (A-Pillar Issues)}  

A필러 상부 강성 부족 및 변형 (Insufficient upper stiffness and deformation in A-pillar)  

A필러 힌지 크랙 발생 (Hinge crack formation in A-pillar)  

A필러와 대시 연결부 찢어짐 (Tearing at the connection between the A-pillar and the dashboard)  

A필러 변형으로 인한 윈드실드 파손 (Windshield damage due to A-pillar deformation)  

A필러 이탈 지연으로 인한 타이어 내측 거동 (Delayed detachment of A-pillar affecting inner tire movement)  

\noindent
\textbf{1.1.2. B필러 문제 (B-Pillar Issues)}  

B필러 하단부 강성 부족으로 인한 변형 과다 (Excessive deformation due to insufficient lower stiffness in B-pillar)  

B필러 용접부 크랙 및 파단 (Cracks and fractures in B-pillar welding area)  

B필러 상단부 꺾임 현상 (Bending at the upper section of the B-pillar)  

B필러와 루프라인 연결부 취약성 (Weak connection between B-pillar and roofline)  

B필러 변형으로 인한 생존 공간 감소 (Reduction in survival space due to B-pillar deformation)  

B필러 부위 접힘 현상 (Folding phenomenon in the B-pillar area)  
B-PLR INR LWR EXTN 누락 (Omission of B-PLR INR LWR EXTN)  

\noindent
\textbf{1.1.3. C필러 및 쿼터패널 문제 (C-Pillar and Quarter Panel Issues)}  

C필러 부위 접힘 현상 (Folding in the C-pillar area)  

쿼터패널 측면 파고듦 현상 (Intrusion on the side of the quarter panel) 

C필러 트림 파손 (C-pillar trim damage)  

쿼터 리테이너 강성 부족 (Insufficient stiffness in quarter retainer)  

\noindent
\textbf{1.2. 도어 구조 관련 문제 (Door Structure Issues)}

\noindent
\textbf{1.2.1. 도어 구조체 문제 (Door Frame Issues)}  

도어 임팩트 빔 마운팅부 파단 및 이탈 (Fracture and detachment of door impact beam mounting)  

도어 빔 브라켓 강도 부족 (Insufficient strength in door beam bracket)  

도어 힌지 마운팅 볼트 파단 및 뽑힘 (Fracture and detachment of door hinge mounting bolts)  

도어 래치 마운팅부 파손 (Breakage in door latch mounting)  

도어 변형으로 인한 침입량 증가 (Increased intrusion due to door deformation)  

임팩트 빔 꺾임 및 마운팅 용접부 파단 (Bending of impact beam and fracture in mounting weld)  

\noindent
\textbf{1.2.2. 도어 부품 및 연결부 문제 (Door Components and Connection Issues)}  

도어 스트라이커 이탈 및 손상 (Detachment and damage of door striker)  

도어 인너 판넬 분리 및 파손 (Separation and breakage of door inner panel)  











\subsection{Pre-written Reports Generation} \label{sec:appendix_prewritten_report_generation}

To generate pre-written expert knowledge, Korean automotive engineers employ the prompt shown in Figure~\ref{fig:expert_report_prompt}, refined through extensive prompt tuning. The generated reports include a structured composition of problem definition, technical analysis, case analysis, improvement suggestions, and relevant references, and are subsequently reviewed by domain experts. The keywords provided in Section~\ref{sec:appendix_keyword} serve as the core problem topics addressed in each report. University-internal Claude service from AWS Bedrock is used for security.

\begin{figure*}[htbp]
    \tiny
    \begin{tcolorbox}[width=1.1\textwidth, colback=white, colframe=black, title=Prompt, sharp corners]
\begin{lstlisting}[
            language=,
            basicstyle=\ttfamily\tiny,
            breaklines=true,
            frame=none
        ]
# You are an expert in writing professional engineering reports in the field of automotive crash safety. Based on the given problem situation and related report data, you must write a technical report that is useful for field engineers.

## Input Format
1. **Problem Situation**: <Problem> {problem} </Problem>
2. **Reference Reports**: <Reports> {reports} </Reports>

## Output Requirements
1. **Title**: A technical title that clearly reflects the problem situation.
2. **Problem Definition**:
   * Detailed description of the problem occurrence.
   * Specification of affected vehicle components.
   * Analysis of the impact on safety.
   * Identification of common problem patterns observed across multiple cases.
3. **Technical Analysis**:
   * Root cause analysis (based on report evidence).
   * Summary of related test results and pattern identification.
   * Inclusion of measured numerical data (if available).
   * Comparative analysis and correlation between multiple tests.
4. **Case Comparison Analysis**:
   * List and categorize all cases where similar issues occurred.
   * Compare and analyze similarities and differences between cases.
   * Compare case results in tabular format (recommended).
5. **Improvement Directions**:
   * Summarize all improvement measures mentioned in the reports.
   * Clearly explain the reasons and expected effects of each improvement.
   * Provide a prioritization of improvement measures.
   * Identify areas requiring additional testing and justify their necessity.
6. **Conclusion**:
   * Summary of key issues.
   * Summary of major findings.
   * Comprehensive expected effects of the proposed improvements.
7. **References**:
   * All information sources must be cited in parentheses right after the corresponding content (e.g., "Due to insufficient upper A-pillar stiffness, the DASH deformation exceeded the target of 100mm by reaching 128mm (C200728A_CV).")
   * Arrange citations appropriately to avoid disrupting readability.

## Key Guidelines
1. **Pattern Identification**: Actively identify and analyze recurring problem patterns across different tests and situations.
2. **Data Integration**: Integrate similar data from multiple reports to provide a more comprehensive analysis.
3. **Importance Evaluation**: Assign higher importance to repeated problems and explicitly indicate this.
4. **Flexible Report Structure**: Use the given structure as a starting point but modify, merge, or add sections as needed based on data and analysis results.
5. Exclude unnecessary introductions or background explanations and focus on core technical content.
6. Do not make assumptions or use general knowledge to fill in missing report data.
7. Provide specific technical information that engineers can use immediately.
8. Ensure all information can be traced back to its source, but citations should not overshadow the main content.
9. Include relevant data, figures, and specific specifications whenever possible.
10. Focus on the methodology applied to solve the problem and the results achieved.
11. Citations should be concise, placed at the end of the sentence or paragraph, and include only the report code in parentheses.

## Citation Examples
* Incorrect: "According to report C200728A_CV, due to insufficient upper A-pillar stiffness..."
* Correct: "Due to insufficient upper A-pillar stiffness, the DASH deformation exceeded the target of 100mm by reaching 128mm (C200728A_CV)."
* Multiple citations: "BrIC is an index developed to assess the risk of brain injury caused by head rotational motion (C200728A_CV, C200730B_NE)."
* Similar issue citation: "The A-pillar deformation issue was consistently observed in three tests (C200728A_CV, C200805B_CV, C200901A_CV), and in all cases, the deformation exceeded the allowable limit by 20-25%."

## Basic Report Structure
```markdown
# Technical Title Based on the Problem Situation
## Problem Definition
- Detailed description of the problem and pattern identification.
- Affected vehicle components.
- Safety impact analysis.
## Technical Analysis
- Root cause analysis.
- Summary of test results.
- Measurement data.
## Case Comparison Analysis
| Test ID   | Issue Description | Measured Value | Standard Value | Deviation | Notes  |
|-----------|-------------------|----------------|----------------|-----------|-----------------|
| TEST-001  | Issue description | 00.0           | 00.0           | 00.0      | Additional info |
| TEST-002  | Issue description | 00.0           | 00.0           | 00.0      | Additional info |
## Improvement Measures and Effects
- Applied solutions.
- Effectiveness measurement results.
- Technical specifications.
## Improvement Directions
- **Priority 1**: [Improvement measure] - [Clear reason and expected effect]
- **Priority 2**: [Improvement measure] - [Clear reason and expected effect]
- **Additional Tests Needed**: [Test details] - [Reason for necessity]
## Conclusion
- Summary of key issues.
- Summary of major findings.
- Comprehensive expected effects of improvements.
\end{lstlisting}
    \end{tcolorbox}
    \caption{Prompt used for generating gold reports. The prompt is originally Korean.}
    \label{fig:expert_report_prompt}
\end{figure*}

\section{On-Demand External Knowledge}\label{sec:appendix_external}
Once the prompt passes the safety filter and deemed to be safe, we use GPT-4o to provide on-demand external knowledge. Specifically, it generates a general-purpose technical background document to help engineers draft formal safety reports. The prompt used is shown in Figure \ref{fig:gpt_external_prompt}. The GPT-generated document is then indexed into the document pool for retrieval.

\begin{figure*}[ht!]
    \tiny
    \begin{tcolorbox}[width=\textwidth, colback=white, colframe=black, title=Prompt, sharp corners]
\begin{lstlisting}[
            language=,
            basicstyle=\ttfamily\tiny,
            breaklines=true,
            frame=none
        ]
    You are an expert in automotive crash safety engineering. 
    You are tasked with generating a general-purpose technical background document **in Korean** that can assist engineers in drafting formal safety reports. 
    Only a high-level keyword describing a common engineering issue is provided. 
    Based on your expert knowledge, please provide background information and reasonable technical discussion, 
    without fabricating any specific case data, test results, or numeric values that are not grounded in well-known general principles.
    
    Your output should include the following sections:
    1. **Title**: A concise technical title derived from the given keyword.
    2. **Issue Overview**:
       - Describe the typical nature of this issue in automotive safety engineering.
       - Mention which vehicle components are generally involved.
       - Explain how this issue may impact structural integrity or passenger safety.
    3. **Common Engineering Considerations**:
       - Discuss known design challenges, structural constraints, or typical causes.
       - Do NOT include fabricated test results or data.
       - Only refer to general engineering principles or commonly reported concerns in literature.
    4. **Improvement Suggestions (General)**:
       - Suggest generic design or material improvements.
       - Clarify expected benefits and rationale (without specific test results).
    5. **Conclusion**:
       - Summarize key considerations.
       - Emphasize that this document provides only general technical guidance, not case-specific findings.
    
    Important:
    - Do NOT fabricate test results, measured data, or specific incident cases.
    - Only refer to general trends, common engineering knowledge, and design principles.
    - This is intended to serve as a general technical reference, not a case analysis.


    keyword: {keyword}
\end{lstlisting}
    \end{tcolorbox}
    \caption{Prompt used for generating on-demand external knowledge using GPT-4o.}
    \label{fig:gpt_external_prompt}
\end{figure*}

\section{Generation}\label{sec:appendix_qwen_train}

We fine-tune Qwen-2.5-14B-Instruct \citep{qwen} using a Korean question-answering dataset focused on the automobile engineering domain. Full fine-tuning is conducted for the 14B model with a batch size of 2, gradient accumulation steps of 64, a learning rate of 2e-5, and 3 training epochs. 3 $\times$ 80GB H100 GPUs are used for the training. Figure \ref{fig:prompt_generation} is the prompt used for report generation of Qwen model.

\begin{figure*}[ht!]
    \tiny
    \begin{tcolorbox}[width=\textwidth, colback=white, colframe=black, title=Prompt, sharp corners]
\begin{lstlisting}[
            language=,
            basicstyle=\ttfamily\tiny,
            breaklines=true,
            frame=none
        ]
You are an expert in writing professional engineering reports in the field of automotive crash safety. Based on the given problem description and related report data, you are required to generate a technical report that provides practical insights for field engineers.

## Output Requirements
1. **Title**: A technical title that clearly reflects the problem.
2. **Problem Definition**:
   * Describe the situation in which the issue occurred in detail.
   * Specify the affected vehicle components.
   * Analyze the impact on safety.
   * Identify common patterns observed across multiple cases.
3. **Technical Analysis**:
   * Analyze the root causes.
   * Summarize related test results and identify patterns.
   * Include measured numerical data.
   * Compare test results and analyze correlations.
4. **Case Comparison Analysis**:
   * List and categorize all cases where similar issues occurred.
   * Compare similarities and differences across cases.
   * Present comparison of results using tables (recommended).
5. **Improvement Directions**:
   * Summarize all improvement actions mentioned in the reports.
   * Clearly explain the rationale and expected effects of each action.
   * Prioritize the improvement directions.
   * Identify areas that require additional testing and justify their necessity.
6. **Conclusion**:
   * Summarize the key problems.
   * Recap the main findings.
   * Synthesize the expected benefits of the proposed improvements.
7. **References**:
   * Cite all sources immediately after the corresponding information using parentheses.
     (e.g., "Due to insufficient upper stiffness of the A-pillar, the DASH deformation exceeded the target value at 128mm compared to the target of 100mm (C200728A_CV).")
   * Ensure references are properly placed without compromising readability.

Ensure that the report is written in a professional and technically accurate manner according to the guidelines above. All conclusions and improvement directions must be based strictly on evidence extracted from the provided reports. Source citations should be placed directly after each statement or piece of information to ensure clarity and traceability.

Pay particular attention to identifying recurring patterns observed across multiple tests or cases and include them in the report. This is critical for understanding the systematic nature of the problem and for proposing effective solutions.

The structure and content of the report must be optimized so that engineers can clearly understand the issue and implement effective solutions. Providing data-driven insights and practical engineering perspectives is key. The conclusion should synthesize the report's core findings and offer clear directions for future development and implementation.

\end{lstlisting}
    \end{tcolorbox}
    \caption{Prompt used for report generation of Qwen model based on the retrieved documents. The prompt is originally Korean.}
    \label{fig:prompt_generation}
\end{figure*}

\section{Evaluation}

\subsection{LLM-based Evaluation} \label{sec:appendix_llm_judge_prompt}

Figure \ref{fig:gpt_eval_prompt} is the prompt that we use for LLM-based evaluation.

\begin{figure*}[ht!]
    \tiny
    \begin{tcolorbox}[width=\textwidth, colback=white, colframe=black, title=Prompt, sharp corners]
\begin{lstlisting}[
            language=,
            basicstyle=\ttfamily\tiny,
            breaklines=true,
            frame=none
        ]
        You are a strict evaluator of AI-generated text for automobile engineering questions.
        You must judge two responses, A and B, on the following metrics:
        1) Correctness: factual correctness based on the question
        2) Richness: level of detail and completeness
        3) Helpfulness: clarity, directness, and overall usefulness
        Then decide who wins each metric (A/B/tie), and also decide the final overall winner.
        Output format strictly:
        CorrectnessWINNER: [A/B/tie]
        RichnessWINNER: [A/B/tie]
        HelpfulnessWINNER: [A/B/tie]
        OVERALLWINNER: [A/B/tie]
        No explanations.


        Question: {question}
        Response A: {answer_a}
        Response B: {answer_b}
        Please provide the winners in the specified format WITHOUT any numerical scores.
\end{lstlisting}
    \end{tcolorbox}
    \caption{Prompt used for LLM-based Evaluation. Pairwise evaluation between Traditional RAG and SecMulti-RAG is conducted using Claude 3.7 Sonnet.}
    \label{fig:gpt_eval_prompt}
\end{figure*}

\subsection{Human and LLM evaluation Agreement} \label{sec:appendix_agreement} 
\begin{figure*}[htbp]
  \centering
  \includegraphics[width=1.2\columnwidth]{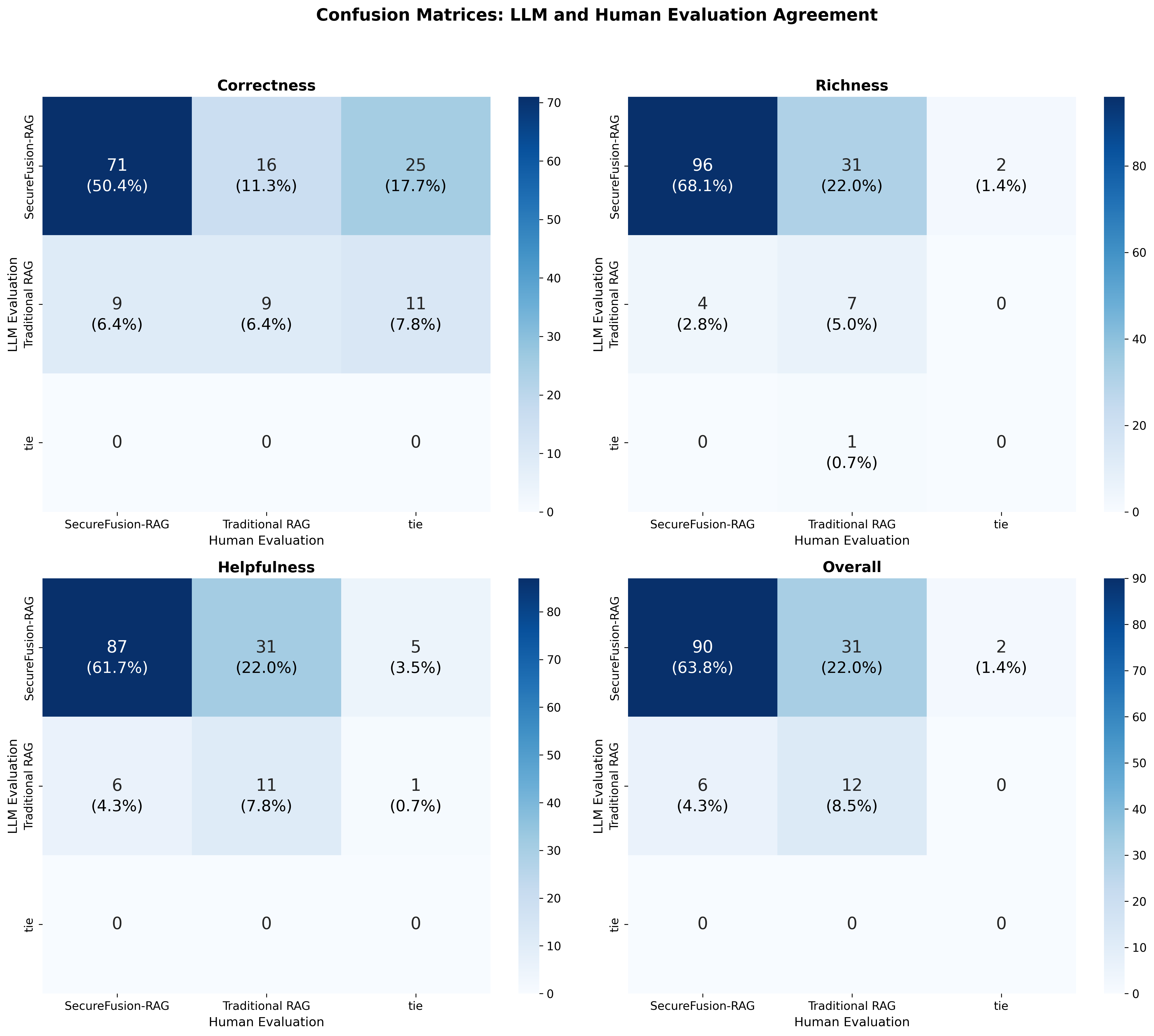}
  \caption{Confusion matrices showing the agreement between LLM and human evaluations. Most counts lie on the diagonal cells, indicating consistent agreement between both evaluators.}
  \label{fig:confusion_matrix}
\end{figure*}

Figure \ref{fig:confusion_matrix} demonstrates confusion matrices showing the agreement between LLM and human evaluations. Each cell contains the count and percentage of cases where Claude (vertical axis) and human evaluators (horizontal axis) made specific judgments about system preference. Darker blue indicates higher frequency. The diagonal cells represent agreement between both evaluators, while off-diagonal cells show disagreement patterns. A large portion of the counts lies along the diagonal, indicating a substantial level of agreement between the two evaluators. Both evaluators show preference for the SecMulti-RAG's generations especially in `Richness' and `Overall' metrics.

Table \ref{tab:agreement_rate} reports the agreement rates and Gwet’s AC1 scores. Gwet’s AC1 is reported instead of Cohen’s Kappa due to the class imbalance in the evaluation results, where SecMulti-RAG is consistently preferred over Traditional RAG. Notably, correctness shows relatively lower agreement, largely due to the LLM’s tendency to avoid assigning `tie' labels, which are more frequently used by human evaluators.

\begin{table}[htbp]
\centering
\begin{tabular}{lcc}
\toprule
\textbf{Metric} & \textbf{Agreement (\%)} & \textbf{Gwet's AC1} \\
\midrule
\textbf{Correctness} & 56.74\% & 0.4295 \\
\textbf{Richness}    & 73.05\% & 0.6812 \\
\textbf{Helpfulness} & 69.50\% & 0.6264 \\
\textbf{Overall}     & 72.34\% & 0.6647 \\
\bottomrule
\end{tabular}
\caption{Agreement between Claude and Human Evaluation Results}
\label{tab:agreement_rate}
\end{table}

\subsection{Task Scalability} \label{sec:appendix_qa} 

\begin{figure*}[ht]
  \centering
  \includegraphics[width=\columnwidth]{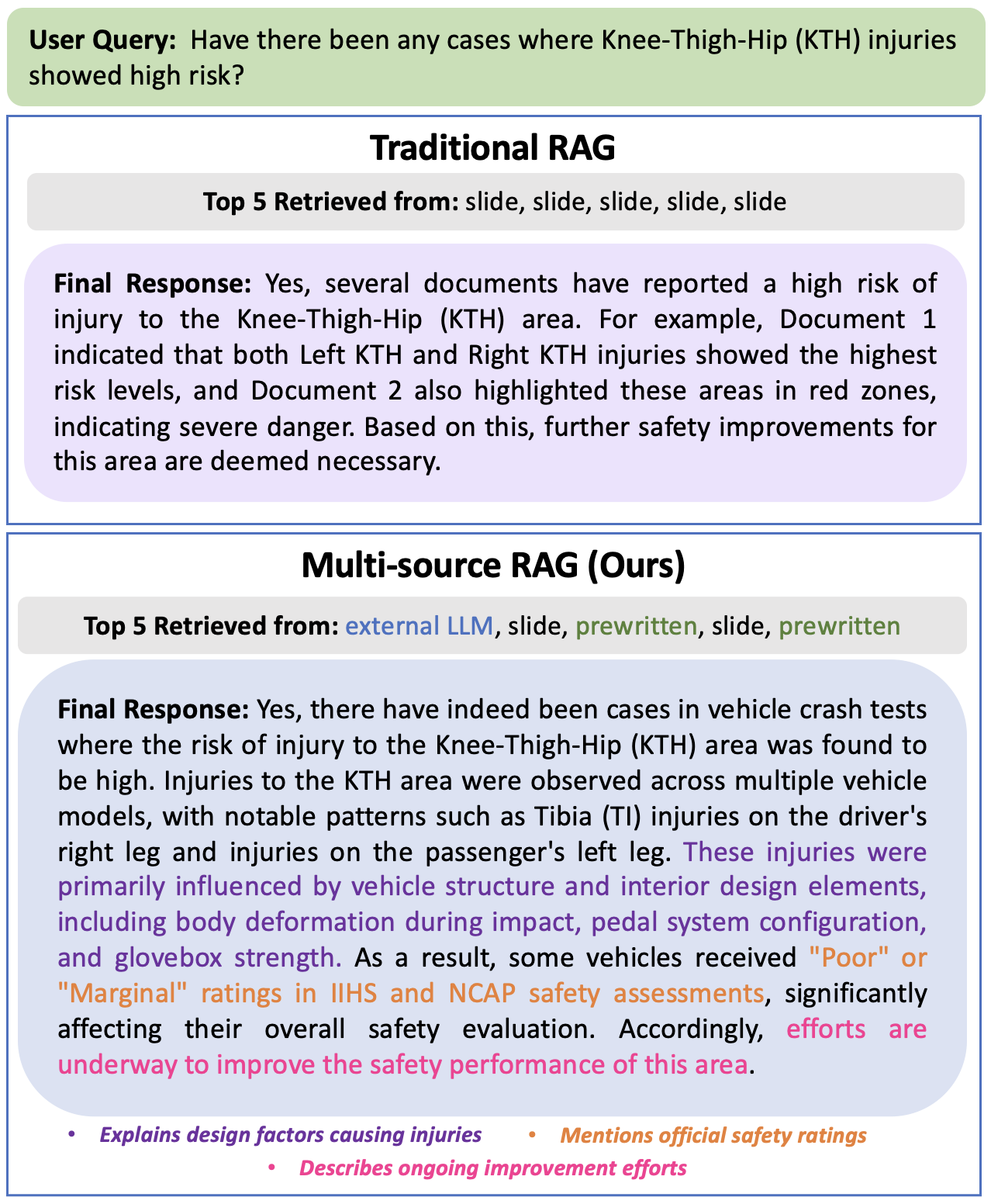}
  \caption{Comparison between Traditional RAG and SecMulti-RAG in QA task (translated from Korean)}
  \label{fig:example_rag}
\end{figure*}

In this study, we primarily evaluate our RAG framework on the report generation task. However, in practice, the framework is scalable to various tasks and domains. We conducted a preliminary test with a few engineering questions in the automotive domain, as shown in Figure~\ref{fig:example_rag}. The SecMulti-RAG responses include specific injury types, structural causes, and implications for official safety assessments, demonstrating greater richness and helpfulness compared to the Traditional RAG responses.

\end{document}